\documentclass[runningheads]{llncs}

\usepackage{eccv}

\usepackage{eccvabbrv}

\usepackage{graphicx}
\usepackage{booktabs}

\usepackage[accsupp]{axessibility}  

\usepackage{hyperref}
\usepackage{marvosym}

\usepackage{orcidlink}

\begin{document}

\title{T-CorresNet: Template Guided 3D Point Cloud Completion with Correspondence Pooling Query Generation Strategy} 

\titlerunning{T-CorresNet: Template Guided 3D Point Cloud Completion}

\author{Fan Duan\orcidlink{0009-0004-1874-9201} \and
Jiahao Yu\and
Li Chen\textsuperscript{(\Letter)}}

\authorrunning{F. Duan et al.}

\institute{School of Software, BNRist, Tsinghua University, Beijing, China\\
\email{chenlee@tsinghua.edu.cn}}

\maketitle

\begin{abstract}
  Point clouds are commonly used in various practical applications such as autonomous driving and the manufacturing industry. However, these point clouds often suffer from incompleteness due to limited perspectives, scanner resolution and occlusion. Therefore the prediction of missing parts performs a crucial task. In this paper, we propose a novel method for point cloud completion. We utilize a spherical template to guide the generation of the coarse complete template and generate the dynamic query tokens through a correspondence pooling (Corres-Pooling) query generator. Specifically, we first generate the coarse complete template by embedding a Gaussian spherical template into the partial input and transforming the template to best match the input. Then we use the Corres-Pooling query generator to refine the coarse template and generate dynamic query tokens which could be used to predict the complete point proxies. Finally, we generate the complete point cloud with a FoldingNet following the coarse-to-fine paradigm, according to the fine template and the predicted point proxies. Experimental results demonstrate that our T-CorresNet outperforms the state-of-the-art methods on several benchmarks. Our Codes are available at \url{https://github.com/df-boy/T-CorresNet}.
\end{abstract}

\section{Introduction}
\label{sec:intro}
Due to the rapid development of scanner technology, 3D point cloud models are extensively employed across various domains. These models have become indispensable elements in fields ranging from autonomous driving to manufacturing industry. 
However, owing to reasons such as limited perspectives, finite scanner resolution, and object occlusion, the point clouds obtained from scanners are typically sparse and incomplete. This poses obstacles to their direct use in reality. Consequently, point cloud completion has gradually emerged as an increasingly important task in the field of 3D computer vision\cite{rusu2008towards, liang2018deep}.

Over the past few years, many deep learning-based methods have been applied to tackle the task of point cloud completion. On account of the success of certain prior in image completion tasks, many early methods like \cite{dai2017shape, han2017high, liu2019relation, wang2017shape, stutz2018learning} tend to transfer some mature approaches to 3D point cloud completion, resulting in an exponential increase in computational complexity. PointNet \cite{qi2017pointnet} and PointNet++ \cite{qi2017pointnet++} pioneer the paradigm of directly processing 3D coordinate streams, leading to the emergence of an entirely new pipeline of point cloud networks, namely encoder-decoder architecture. A large number of methods based on this framework have been proposed \cite{huang2020pf, mandikal2019dense, tchapmi2019topnet, yuan2018pcn}, which typically extract a coarse-grained feature vector from the incomplete input and decode it into a complete point cloud model during the decoding phase. However, these methods tend to lose fine-grained details during the encoding phase due to the pooling operations, thus affecting the local geometric details of the completion results. Additionally, some cascaded refinement methods gradually refine point clouds by predicting offsets of point cloud distributions. Nevertheless, these methods also mainly rely on the global feature resulting in geometric deficiencies in the fine-grained detail of the reconstructed point clouds.

Thanks to the development of Transformers \cite{vaswani2017attention}, the ability of natural language processing (NLP) models to model long-range sequential dependencies has been greatly improved. PoinTr \cite{yu2021pointr}, AdaPoinTr \cite{yu2023adapointr} and ProxyFormer \cite{li2023proxyformer} combine the encoder-decoder architecture with the Transformers, transforming the point cloud completion task into a set-to-set translation problem. With the local point proxies translation, they effectively enhance the accuracy and geometric details of generated results. SeedFormer \cite{zhou2022seedformer} introduces the Patch Seeds to preserve the local information to maintain fine-grained geometric details and utilize the Transformer-like up-sample module to complete the point cloud in a coarse-to-fine manner. Nevertheless, dependencies between text sequences in NLP problems are usually built upon the model's understanding of language continuity and semantics which will bring up several deficiencies: 

(i) The lack of global semantic priors in the 3D space of these methods would increase the complexity of the solution space for set-to-set translation problems, thereby affecting the accuracy of the final generated point clouds;

(ii) Existing methods typically tend to reuse the transformer decoder part of the Transformer paradigm, predicting the fine-grained point-level features of the complete point cloud with the features of the input point cloud and the positional information. However, the dependency between the input partial point cloud proxies and the missing parts can be greatly complex due to different missing scenarios of the models. 

Since the point cloud completion is similar to the shape generation task, the commonly used deformable template which can serve as the global prior might be the answer to the first issue. And for the second problem, new query tokens and value vectors that can be suitable for the 3D scene might be able to predict the complete point proxies in various incomplete situations.

In order to address these issues, we propose a novel point cloud completion method that fetches the global semantic priors with a gaussian template and models the dependency between the fused template and complete point cloud through a Corres-Pooling query generator. Overall, our T-CorresNet combines the transformer encoder-decoder architecture, following the coarse-to-fine generation paradigm. Our encoder module mainly consists of two main phases: (i) extract the point proxies of the input point cloud and fuse the standard Gaussian spherical template into the point proxies of the partial input; (ii) model the long-distance dependency with a geometry-aware transformer encoder \cite{yu2021pointr} and generate a coarse complete template with the fused point proxies. Unlike existing methods, we use a spherical template to enlarge the incomplete input space and guide the generation of the coarse complete point cloud at the same time, improving the semantic knowledge of the global shape prior. 
Then we use the Corres-Pooling query generator to refine the coarse template and generate the dynamic query tokens. Specifically, we combine the coarse template with the sampled partial input to construct a Corres-Pool. And then through the correspondence attention(Corres Attention) mechanism, we remove a certain number of points with the highest relevance to the input from the coarse template to maintain the raw information of the partial input. Through a voting network, we select the top $N_0$ ($N_0$ will be introduced later) points with highest scores in the pool to compose the dynamic query tokens and the fine template. Finally, with the query tokens and spherical value vectors, we can predict the point proxies through a transformer decoder which will be fed into a FoldingNet\cite{yang2018foldingnet} to generate the complete result. 

Our main contributions could be summarized as follows:

(1)We propose a novel T-CorresNet that utilizes a spherical template to guide the generation of the coarse complete point cloud, granting the network a better global understanding of the model shape;

(2)We design a correspondence-pooling(Corres-Pooling) query generator that can generate the dynamic query tokens which can maintain the raw information from the partial input and strengthen the ability to predict the point proxies;

(3)Qualitative and quantitative experiments on several commonly used benchmarks have demonstrated that our T-CorresNet outperforms the state-of-the-art methods.

\section{Related Work}
\subsubsection{Voxel-based Shape Completion.} 
The early methods for 3D shape completion often utilize the structured voxel grids to represent 3D objects\cite{dai2017shape, han2017high, stutz2018learning}. 3D convolutions, derived from 2D fields, are thus naturally and effectively applied to various 3D shape reconstruction and completion tasks based on these representations. Based on the strong 3D convolutions, several methods \cite{wu20153d, dai2017shape, han2017high} have achieved certain progress in the 3D shape completion tasks. However, these methods tend to consume great computational complexity owing to the voxel resolution. 

\subsubsection{Point Cloud Completion.} 
As an unstructured representation of 3D models, point clouds are increasingly favored by researchers in the task of 3D shape completion due to their low memory consumption and ability to represent fine details. However, the 3D convolutions are not that suitable for the unstructured point clouds. PointNet \cite{qi2017pointnet} proposes a pioneering design to directly process 3D coordinate streams by the MLPs and fetch the aggregated features with pooling operations. Based on this upstream design, various methods \cite{huang2020pf, mandikal2019dense, tchapmi2019topnet, yuan2018pcn} have been proposed in terms of the point cloud completion task. PCN \cite{yuan2018pcn} is the first deep neural network which predicts the global feature of the complete model with an encoder-decoder architecture and generates the final complete point cloud with a FoldingNet\cite{yang2018foldingnet}. Afterwards, many similar methods extend the modules of PCN in different perspectives, leading to improvements in both the accuracy and robustness of point cloud completion. SnowflakeNet \cite{xiang2021snowflakenet} proposes to simulate the process of point cloud completion as the growth of snowflakes, achieving the point cloud generation by gradually expanding and spreading parent feature points in 3D space. PMP-Net\cite{wen2021pmp} and PMP-Net++ \cite{wen2022pmp} propose to mimic the earth mover, learning the multi-step point moving paths to generate the complete point cloud by limiting the total point moving distance. Nevertheless, these methods often tend to lose certain fine-grained details during the expanding phase. With the development of Transformers\cite{vaswani2017attention} in MLP, the ability of Transformers to model the long-range dependency between the sequences gradually arouses the interest of researchers from other fields. PoinTr \cite{yu2021pointr}, ProxyFormer \cite{li2023proxyformer} and AdaPoinTr \cite{yu2023adapointr} propose to combine the encoder-decoder architecture with the Transformers, transforming the point cloud completion task into a set-to-set translation problem and using the local point proxies to maintain the fine-grained details. SeedFormer \cite{zhou2022seedformer} introduces Patch Seeds to represent the point clouds and recover fine details in the up-sample transformer layers. However, dependencies between text sequences are often built upon the global understanding of semantics continuity. Thus the lack of global semantic priors would increase the complexity of the solution space for the completion task and affect the accuracy of the completion results. At the same time, reusing the transformer decoder without generating the query and value tokens that are tailored for the 3D scene could do more harm than good to the strong ability of the Transformers.

\subsubsection{Transformers.} Transformers \cite{vaswani2017attention} are first used in Natural Language Processing (NLP) domain, with an extremely strong ability to model the dependencies between long sequences through the attention mechanisms. These attention mechanisms with the encoder-decoder architecture could help process the long sequence and model the attention map between the query tokens and the value vectors at the same time, which could be key to enhancing the feature communications between the encoder and the decoder. The past few years have witnessed the widespread applications of Transformer-based modes in computer vision\cite{parmar2018image, yu2022point}. PoinTr\cite{yu2021pointr} first attempt to apply the Transformers into 3D point cloud completion task and achieve great success. ProxyFormer\cite{li2023proxyformer}, SeedFormer\cite{zhou2022seedformer} and AdaPoinTr\cite{yu2023adapointr} respectively extend it in different perspectives to gain finer-grained details. However, there still exist certain deficiencies in these methods as we discussed before. In this work, we further explore the Transformer module, introducing the spherical template and the Corres-Pooling query generator. This enhancement enables our T-CorresNet to achieve a more comprehensive global understanding of the complete point cloud, seamlessly integrating the Transformer module into the point cloud completion task.

\section{Method}
\begin{figure}[tb]
  \centering
  \includegraphics[width=1.0\linewidth]{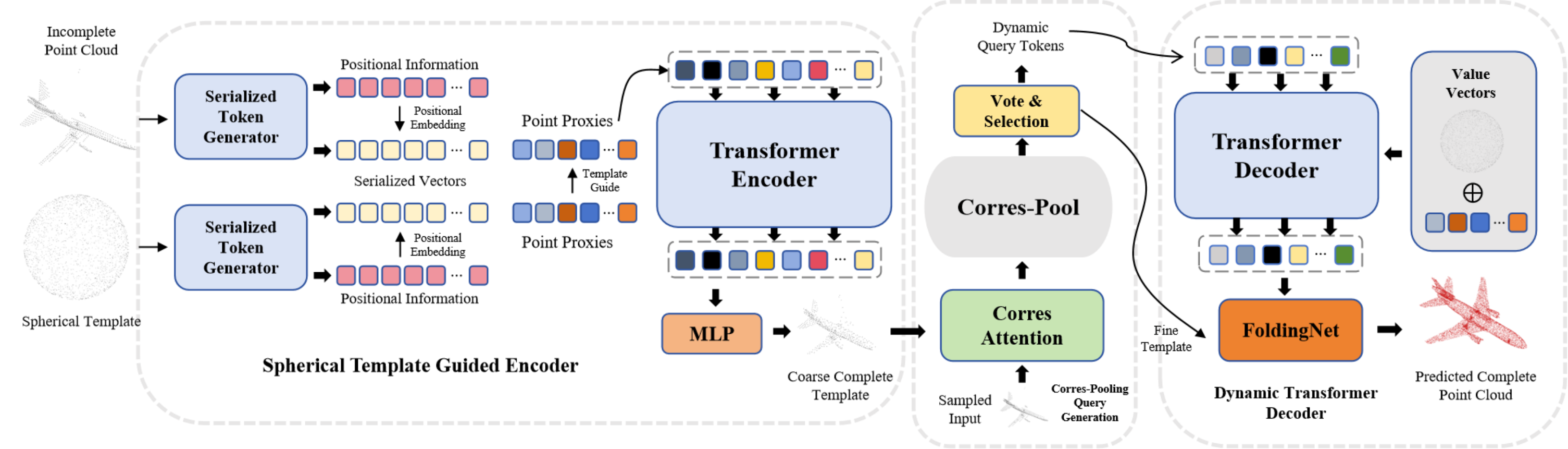}
  \caption{
  An overview of the proposed T-CorresNet. We combine the transformer encoder-decoder architecture with the spherical template, following the coarse-to-fine generation paradigm. Our encoder module mainly predicts the global point proxies and generates the coarse complete template model under the guide of the spherical prior. Then through the Corres-Pooling query generator, we fetch the dynamic query tokens along with the value vectors, which could predict the accurate complete point proxies by the transformer decoder module. The FoldingNet could generate the complete results based on the fine template and the predicted point proxies. 
  }
  \label{fig:overview}
\end{figure}

In this section, we will explain the details of T-CorresNet. The overall architecture of our design is illustrated in \cref{fig:overview}. We will show the details of our encoder module in \cref{sec:encoder} to introduce the spherical template guided encoder, and then we will present our Corres-Pooling query generation and the decoder module in \cref{sec:corres-pooling} and \cref{sec:decoder}. In the end, we show the loss function used in T-CorresNet in \cref{sec:loss}.
\subsection{Spherical Template Guided Encoder}
\label{sec:encoder}
The spherical template guided encoder module is aimed to extract the point proxies from the input point cloud and generate the coarse complete template under the guidance of the standard spherical template. 
\subsubsection{Gaussian Spherical Template}
Point cloud completion tasks typically require establishing a mapping from existing partial input to complete point cloud. From a serialized perspective, this involves constructing a mapping from the serialized token of the existing input to that of the complete model. Currently, most existing methods tend to impose mandatory mapping constraints, using the ground truth as supervision. Actually, the point cloud completion task is another form of shape generation. In SP-GAN\cite{li2021sp}, the standard unit sphere is used for shape generation and this design achieves great success on the shape generation benchmarks compared with other methods. Moreover, SP-GAN explores the ability of other shapes like the unit cube, and the results turns out that the unit sphere is the best for the shape generation. In other words, the sphere as a template demonstrates extremely strong flexibility. Therefore, we propose to embed the partial input with a standard spherical template, thereby expanding the input space. This allows different types of missing scenarios to achieve higher consistency after template embedding and facilitates the learning of the mapping from the input to the complete result. The generation of the standard spherical template can be formulated as follows. First we sample $N$ points $S = \{s_1, s_2, ..., s_n\}$ from the standard Gaussian distribution:
\begin{align}
  s_i \sim N(0, 1).
\end{align}
Then we normalize these points to generate the standard Gaussian spherical template:
\begin{align}
s_i = \frac{s_i}{||s_i||_2}.
\end{align}
With this spherical template, we could use it to guide the generation of the coarse complete template in the subsequent modules.

\subsubsection{Serialized Token Generator}
In order to fully apply the strong ability of Transformer module into our point cloud completion task to model the sequence-to-sequence dependency, we first use the serialized token generator to transform the input point cloud into a set of abstract feature vectors, namely point proxies, which can represent the local space of the input point cloud. Specifically, denote the input incomplete point cloud as $\mathcal{P} = \{p_1, p_2, p_3, ...,p_n\} \in \mathcal{R} ^{N \times 3}$, where $N$ is the number of the points in the input. Inspired by PoinTr\cite{yu2021pointr}, we use a DGCNN\cite{wang2019dynamic}-like network to extract the local feature from the downsampled input model, which can be formulated as:
\begin{equation}
  \mathcal{F'} = \text{DGCNN}(\textit{fps}(\mathcal{P})).
\end{equation}
where $\textit{fps}$ is the farthest point sampling that can fetch the key local points of the point cloud, and $\text{DGCNN}$ represents the dynamic graph convolution. Through the serialized token generator, we can fetch the local point features $\mathcal{F'} \in \mathcal{R} ^{M \times C}$ and the sampled point cloud coordinates $\mathcal{P'} \in \mathcal{R} ^ {M \times 3}$. In order to generate the point proxies which can definitely represent the local regions of the key points, we need to embed the positional information into the local point features:
\begin{align}
    \mathcal{F} =\varphi_{1}(\mathcal{F'}) + \varphi_{2}(\mathcal{P'}).
\end{align}
where $\varphi_1$ and $\varphi_2$ are independent MLPs to map the point features and positional coordinates into proxy space respectively. In this way, we could fetch the point proxies of the input point cloud that represent the complete local information of key points in the input. The 3D input is thus transformed into 1D serialized tokens, laying the foundation for modelling a sequence-to-sequence correspondence in the subsequent encoder.
In the same manner, the 3D standard spherical template, which has been introduced above, could also be transformed into 1D serialized tokens.
\subsubsection{Graph Transformer Encoder Block} 
Compared to some traditional neural networks applied in the 3D domain, the main drawback of directly applying the vanilla Transformer to the 3D domain is the lack of inference bias towards the topology of 3D models. Therefore, we follow the geometry-aware Transformer block\cite{yu2021pointr}, using $k\text{NN}$ together with the self-attention mechanism to model the feature relations between the points.
For the query point $p_{query}$, we obtain the $k$ nearest points based on $k\text{NN}$ in terms of feature level, denoted as the point set $N(p_{query})$. And then we concatenate this feature relations with the sequential feature computed by the self-attention mechanism. This process can be formulated as:
\begin{equation}
    f(p_{query}) = \text{MLP}(\mathcal{F}_{p_{query}}, \mathcal{F}_{p_j} - \mathcal{F}_{p_{query}}) \bigoplus f'(p_{query}), \forall p_j \in N(p_{query}).
\end{equation}
where $f'(p_{query})$ represents the sequential features extracted from the multi-head self-attention mechanism in the vanilla Transformer block, $\bigoplus$ stands for the channel-level concatenation and $\mathcal{F}$ means the point proxies generated in the serialized token generator. Lastly, we map this feature to the original space to generate the transformed point proxies which will be converted into the coarse complete template through a single MLP. The point proxies of the standard spherical template could guide the generation through the layers of transformer block, which can be formulated as below:
\begin{equation}
    \mathcal{F}_{input} = \text{E}(\mathcal{F}_{input} + \mathcal{F}_{t}).
\end{equation}
where E represents the transformer encoder block, $\mathcal{F}_{input}$ and $\mathcal{F}_t$ represent the point proxies of the partial input and the spherical template respectively. In this manner, the spherical template can be embedded into the partial input in the transformer layers and enlarge the input space, which could guide the generation and simplify the mapping to solution space. Lastly, we use a single MLP and max pooling operation to generate the coarse complete template, which can be formulated as follows:
\begin{align}
    \mathcal{P}_{coarse\ Template} = \text{MLP}(\mathcal{M}(\mathcal{F}_{input})).
\end{align}
where $\mathcal{M}$ is the max pooling operation.
\subsection{Corres-Pooling Query Generation}
\label{sec:corres-pooling}
In the aforementioned encoder modules, we expand the input space by guiding the model generation through embedding a spherical template into the partial input and generate the coarse complete template. However, in this process, due to the randomness of the Gaussian template sphere, we inevitably introduce some noise into the coarse template. Therefore, in order to denoise and refine the coarse complete template, we propose the Corres-Pooling query generation strategy, which enables our model to obtain a finer template and generate dynamic query tokens for predicting the point proxies of the complete point cloud.
\begin{figure}[tb]
  \centering
  \includegraphics[width=1.0\linewidth]{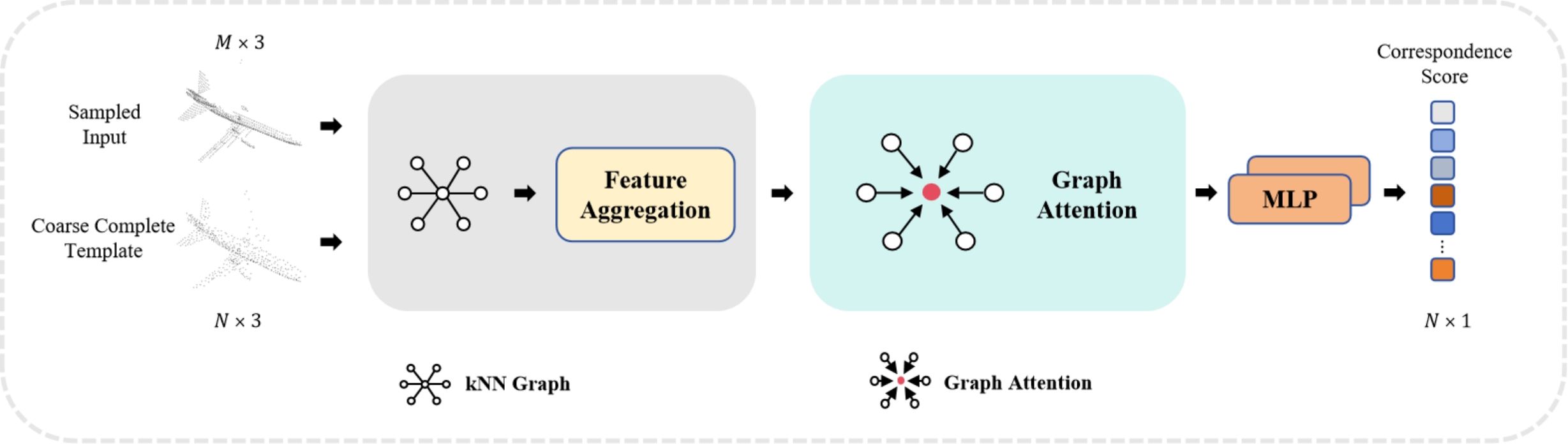}
  \caption{
  The pipeline of Corres Attention module. 
  }
  \label{fig:corresattention}
\end{figure}

Since locating the noise in the coarse template without any supervision is hard to achieve, the only controllable part of the template is the point set that represents the partial input. Therefore we propose the Corres Attention module to identify those points which have high similarity with the input and substitute them with the partial input in the subsequent operations. The details of Corres Attention module are shown in \cref{fig:corresattention}.

Corres Attention module first constructs a $k\text{NN}$ graph between the coarse complete template and the sampled input, then aggregates the features according to this graph. To model the long-range dependency and utilize the topology information, we use a graph attention module here. The main difference between our graph attention module and the vanilla attention mechanism is that, for the query point $p_i$ in the template, the key and value vectors are both replaced with the aggregated features, which are computed from the $k$ nearest points in the sampled input with $k\text{NN}$ and linear layers. Then we use a two-layer MLP to predict the correspondence score which represents the similarity of each point in the template to the partial input.

We replace the top $k$ points in the template with the sampled $k$ points from the partial input according to the aforementioned correspondence scores. Then these points form the candidate points pool, namely Corres-Pool, from where we will generate the fine template the dynamic query tokens with the single voting mechanism. Specifically, we predict the scores of each point in the Corres-Pool using several MLPs and select a certain number of points to form the fine template. The generation of the dynamic query tokens can be formulated as:
\begin{align}
    \mathcal{Q} = P_{fine Template} \bigoplus \mathcal{M} (\mathcal{F}_{input}).
\end{align}
where the $\mathcal{M}(\mathcal{F}_{input})$ is the global feature generated by the max pooling operation towards the output of the transformer encoder block and the $\bigoplus$ is the channel-wise concatenation.

\subsection{Dynamic Transformer Decoder}
\label{sec:decoder}
The key idea of our dynamic transformer decoder is similar to that in the Transformer encoder block, using $k\text{NN}$ together with the self-attention and cross-attention mechanisms to model the feature relations between the points. It is worth mentioning that not only are our query tokens dynamically generated, but our value vectors are also well designed to fully enhance the decoder. The generation of the value vectors can be formulated as:
\begin{align}
    \mathcal{V} = \mathcal{F}_{input} \bigoplus \text{Random} (S).
\end{align}
where $\mathcal{F}_{input}$ means the output of the transformer encoder block and $\text{Random}(S)$ means a new standard sphere as mentioned above. The standard sphere could make the value vectors more suitable to the query tokens in the Transformer decoder block. Then with the global point proxies and the fine template, we utilize a FoldingNet\cite{yang2018foldingnet} to recover the detailed local shapes of the fine template. Specifically, for each point $p_i$ with the predicted point proxy $\mathcal{\hat{F}}_i$:
\begin{align}
    \mathcal{N}(p_i) = f(\mathcal{\hat{F}}_i) + p_i.
\end{align}
where $\mathcal{N}(p_i)$ is the neighbour points centered at $p_i$ and $f(\mathcal{\hat{F}}_i)$ is the bias value predicted by FoldingNet. In this way, we generate the predicted complete point cloud with the fine template and the global point proxies predicted by the dynamic Transformer decoder block in a coarse-to-fine manner. 

\subsection{Loss Functions}
\label{sec:loss}
The learning objectives for point cloud completion need to measure the measure the distance between two unordered point sets, therefore some metrics that compute the distance between two points (i.e. $l_1, l_2$ loss) are not that suitable for this task. We choose Chamfer Distance(CD) as our loss function for its invariance to the unordered points. Given the ground truth complete point cloud $\mathcal{G}$, we use $\mathcal{T}$ and $\mathcal{P}$ to represent the fine template and the predicted complete point cloud respectively. The loss functions can be formulated as:
\begin{align}
    \mathcal{L}_0 &= \frac{1}{n_{\mathcal{T}}} \sum_{t\in \mathcal{T}}\min_{g\in \mathcal{G}} ||t- g|| + \frac{1}{n_{\mathcal{G}}} \sum_{g\in \mathcal{G}}\min_{t\in \mathcal{T}} ||g- t||, \\
    \mathcal{L}_1 &= \frac{1}{n_{\mathcal{P}}} \sum_{p\in \mathcal{P}}\min_{g\in \mathcal{G}} ||p- g|| + \frac{1}{n_{\mathcal{G}}} \sum_{g\in \mathcal{G}}\min_{p\in \mathcal{P}} ||g- p|| .
\end{align}
It is worth mentioning that we directly use the complete point cloud $\mathcal{G}$ to supervise the sparse fine template, which could make the distribution of the fine template similar to that of $\mathcal{G}$. The total training loss is formulated as:
\begin{equation}
    \mathcal{L} = \mathcal{L}_0 + \mathcal{L}_1.
\end{equation}

\section{Experiments and Results}
\subsection{Implementation Details}
T-CorresNet is implemented with PyTorch on one NVIDIA GeForce RTX 3090 GPU. The main hyper-parameters are the $k$s used in $k\text{NN}$, which are set 16. Additionally, the coarse template and the fine template both contain $N_0 = 512$ points, and the query token pool generated by the Corres-Pooling mechanism holds $N_1 = 640$ points which is composed of $N_3 = 256$ points from the coarse template and $N_4 = 384$ points from the sampled partial input. The learning rate is set to 0.0001 and we use the cosine annealing learning rate scheduler\cite{loshchilov2016sgdr} to improve the generalization ability of our model.
\subsection{Benchmarks and Metrics}
In order to evaluate our method on real-world 3D objects, we choose several common benchmarks which have various object categories and more viewpoints.
\subsubsection{PCN Benchmark} The PCN dataset\cite{yuan2018pcn} is a subset of ShapeNet\cite{chang2015shapenet}, including 30974 3D objects with eight categories. PCN dataset is the most widely used benchmark in the 3D point cloud completion task. We keep our experimental setting the same with PCN for the sake of fairness.
\subsubsection{ShapeNet-55 Benchmark} 
The ShapeNet-55 Benchmark\cite{yu2021pointr} is also extracted from the synthetic ShapeNet\cite{chang2015shapenet}. Compared to PCN, which only has eight object categories, the ShapeNet-55 benchmark contains 55 categories of real-world 3D objects, making it more realistic. Our experiments on the ShapeNet-55 benchmark follow the setting of PoinTr\cite{yu2021pointr} to achieve a fair comparison.
\subsubsection{ShapeNet-34 Benchmark} 
ShapeNet-34 benchmark\cite{yu2021pointr} is used to test the generalization ability of the model, which is divided into two parts: 34 seen categories for training and 21 unseen categories for evaluation. The detailed experimental settings of this benchmark are the same as that in PoinTr\cite{yu2021pointr}.

\subsubsection{Evaluation Metrics}
For the PCN benchmark, we sample 2048 points as the partial input and 16384 points as the ground truth, following PCN. For the ShapeNet-55 benchmark and ShapeNet-34 benchmark, we sample 2048, 4096 or 6144 (25\%, 50\%, 75\% of the complete point cloud) points as partial input and 8192 points as the ground truth which represents the easy, median and hard mode respectively, keeping it the same as PoinTr. Similar to the previous methods\cite{yuan2018pcn, yu2021pointr, yu2023adapointr, zhou2022seedformer}, we adopt Chamfer distance and F1 score as the evaluation metrics. Given the predicted point cloud $\mathcal{P}$ and the ground truth $\mathcal{G}$, the calculation of the Chamfer distance can be formulated as:
\begin{align}
    \text{CD}(\mathcal{P}, \mathcal{G}) = \frac{1}{|\mathcal{P}|} \sum_{p\in \mathcal{P}}\min_{g\in \mathcal{G}} ||p- g|| + \frac{1}{|\mathcal{G}|} \sum_{g\in \mathcal{G}}\min_{p\in \mathcal{P}} ||g- p||.
\end{align}
We use CD-$l_1$ with L1-norm for the PCN benchmark and CD-$l_2$ with L2-norm for the ShapeNet-55 and ShapeNet-34 benchmarks respectively. Following \cite{tatarchenko2019single} to adopt F-Score as an evaluation metric, we set the threshold of F-Score to 1\% in our experiments.
\subsection{Results on PCN}
\begin{table}[t]
\caption{Quantitative results on the PCN dataset in terms of CD-$l_1$ $\times$ 1000(lower is better) and F-Score@1\%(higher is better).}
\centering
\begin{tabular}{c|cccccccc|cc}
\toprule[2pt]
Methods & Air & Cab & Car & Cha & Lam & Sof & Tab & Wat & Avg CD & F1 \\
\hline
FoldingNet\cite{yang2018foldingnet} & 9.49 & 15.80 & 12.61 & 15.55 & 16.41 & 15.97 & 13.65 & 14.99 & 14.31 & 0.322 \\
AtlasNet\cite{groueix2018papier} & 6.37 & 11.94 & 10.10 & 12.06 & 12.37 & 12.99 & 10.33 & 10.61 & 10.85 & 0.616 \\
PCN\cite{yuan2018pcn} & 5.50 & 22.70 & 10.63 & 8.70 & 11.00 & 11.34 & 11.68 & 8.59 & 9.64 & 0.695 \\
TopNet\cite{tchapmi2019topnet} & 7.61 & 13.31 & 10.90 & 13.82 & 14.44 & 14.78 & 11.22 & 11.12 & 12.15 & 0.503\\
GRNet\cite{xie2020grnet} & 6.45 & 10.37 & 9.45 & 9.41 & 7.96 & 10.51 & 8.44 & 8.04 & 8.83 & 0.708 \\
PoinTr\cite{yu2021pointr} & 4.75 & 10.47 & 8.68 & 9.39 & 7.75 & 10.93 & 7.78 & 7.29 & 8.38 & 0.745 \\
PMP-Net\cite{wen2021pmp} & 5.65 & 11.24 & 9.64 & 9.51 & 6.95 & 10.83 & 8.72 & 7.25 & 8.73 & - \\
CRN\cite{wang2020cascaded} & 4.79 & 9.97 & 8.31 & 9.49 & 8.94 & 10.69 & 7.81 & 8.05 & 8.51 & - \\
NSFA\cite{zhang2020detail} & 4.76 & 10.18 & 8.63 & 8.53 & 7.03 & 10.53 & 7.35 & 7.48 & 8.06 & - \\
SnowflakeNet\cite{xiang2021snowflakenet} & 4.29 & 9.16 & 8.08 & 7.89 & 6.07 & 9.23 & 6.55 & 6.40 & 7.21 & - \\
LAKeNet\cite{tang2022lake} & 4.17 & 9.78 & 8.56 & 7.45 & 5.88 & 9.39 & 6.43 & 5.98 & 7.23 & - \\
SeedFormer\cite{zhou2022seedformer} & 3.85 & 9.05 & 8.06 & 7.06 & \textbf{5.21} & 8.85 & 6.05 & 5.85 & 6.74 & - \\
ProxyFormer\cite{li2023proxyformer} & 4.01 & 9.01 & 7.88 & 7.11 & 5.35 & 8.77 & 6.03 & 5.98 & 6.77 & - \\
AdaPoinTr\cite{yu2023adapointr} & 3.68 & 8.82 & \textbf{7.47} & 6.85 & 5.47 & 8.35 & \textbf{5.80} & 5.76 & 6.53 & \textbf{0.845} \\
\hline
T-CorresNet & \textbf{3.63} & \textbf{8.79} & 7.48 & \textbf{6.72} & 5.22 & \textbf{8.28} & 5.92 &\textbf{5.70} & \textbf{6.47} & \textbf{0.845} \\
\bottomrule[2pt]
\end{tabular} 
\label{tab:PCN}
\end{table}
\begin{figure}[t]
    \centering
    \includegraphics[width=1.0\linewidth]{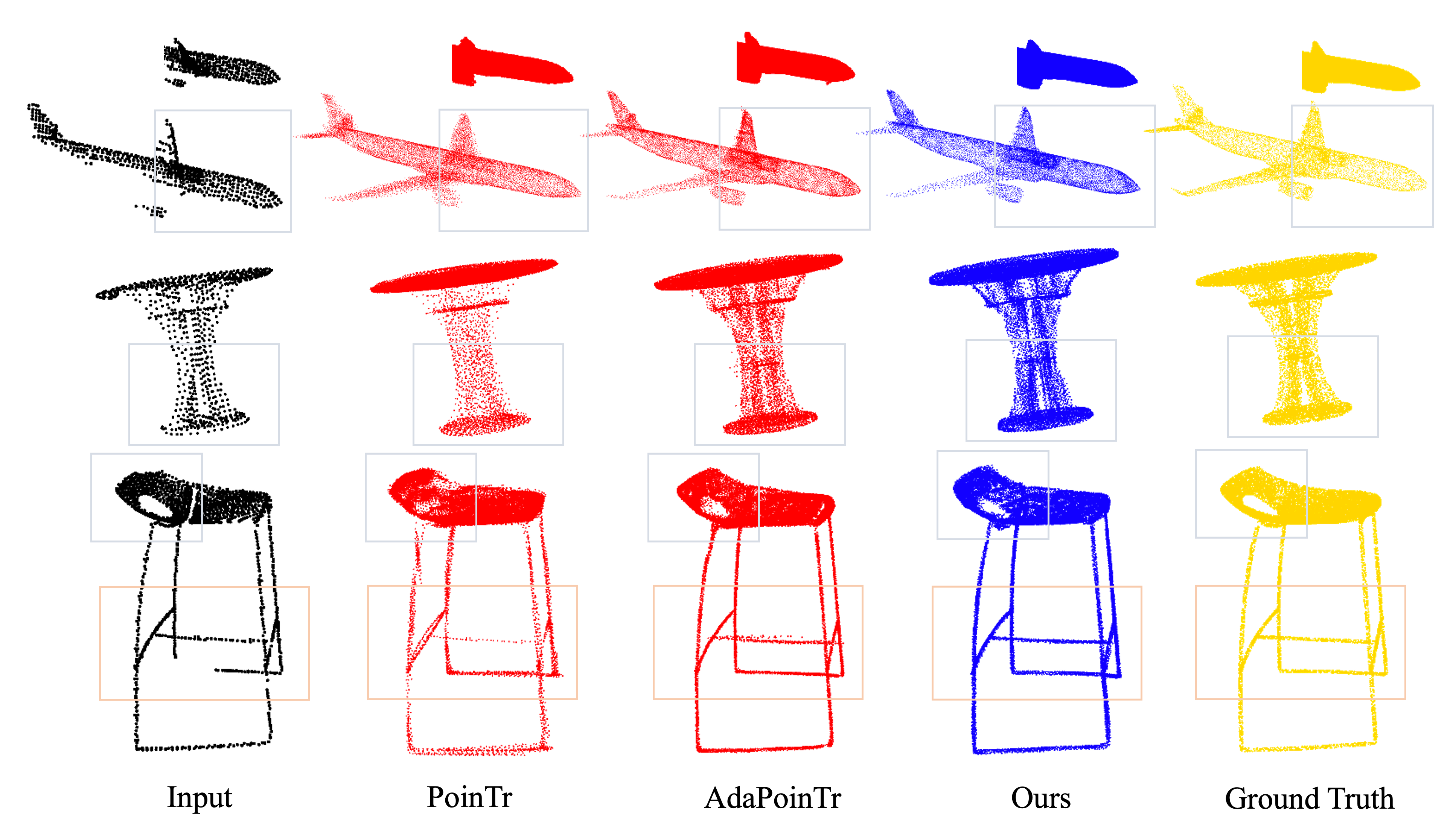}
    \caption{Visualization results on PCN dataset.}
    \label{fig:vis}
\end{figure}
To evaluate our method on the PCN benchmark, we conduct several experiments on the recent methods following previous methods. The quantitative results are shown in \cref{tab:PCN}. The CD-$l_1$ and F-score@1\% metrics are provided in the table. Our T-CorresNet improves the performance on most categories and outperforms the previous state-of-the-art methods. These results demonstrate that our T-CorresNet with the template guide is definitely effective in real-world scenarios. \cref{fig:vis} shows the visualization results from three categories(airplane, table and chair) compared with PoinTr and AdaPoinTr. Overall, there is a significant gap between the completion results from PoinTr and those from the latter two methods. And it can be noticed that our T-CorresNet maintains more faithful details on the shapes. For instance, the wings reconstructed by T-CorresNet better match the shape of the ground truth, while AdaPoinTr tends to generate the wheel shapes even for airplane models without wheels. Additionally, AdaPoinTr tends to fill in the holes for chairs with hole-shaped designs with much noise. In contrast, our model achieves a more realistic completion result with the global template prior in these aspects.
\subsection{Results on ShapeNet-55/34}
\begin{table}[t]
    \caption{Quantitative results on the ShapeNet-55 dataset in terms of CD-$l_2$ $\times$ 1000(lower is better) and F-Score@1\%(higher is better).}
    \centering
    \begin{tabular}{c|cccccc|ccc|cc}
    \toprule[2pt]
        Methods &  Tab & Cha & Air & Bir & Bag & Key & CD-S & CD-M & CD-H & Avg CD & F1 \\
        \hline
        FoldingNet\cite{yang2018foldingnet} & 2.53 & 2.81 & 1.43 & 4.71 & 2.79 & 1.24 & 2.67 & 2.66 & 4.05 & 3.12 & 0.082 \\
        PCN\cite{yuan2018pcn} & 2.13 & 2.29 & 1.02 & 4.50 & 2.86 & 0.89 & 1.94 & 1.96 & 4.08 & 2.66 & 0.133 \\
        TopNet\cite{tchapmi2019topnet} & 2.21 & 2.53 & 1.14 & 4.83 & 2.93 & 0.95 & 2.26 & 2.16 & 4.3 & 2.91 & 0.126 \\
        PFNet\cite{huang2020pf} & 3.95 & 4.24 & 1.81 & 6.21 & 4.96 & 1.29 & 3.83 & 3.87 & 7.97 & 5.22 & 0.339 \\
        GRNet\cite{xie2020grnet} & 1.63 & 1.88 & 1.02 & 2.97 & 2.06 & 0.89 & 1.35 & 1.71 & 2.85 & 1.97 & 0.238 \\
        PoinTr\cite{yu2021pointr} & 0.81 & 0.95 & 0.44 & 1.86 & 0.93 & 0.38 & 0.58 & 0.88 & 1.79 & 1.09 & 0.464 \\
        SnowflakeNet\cite{xiang2021snowflakenet} & 0.98 & 1.12 & 0.54 & 1.93 & 1.08 & 0.48 & 0.70 & 1.06 & 1.96 & 1.24 & 0.398\\
        LAKeNet\cite{tang2022lake} & - & - & - & - & - & - & - & - & - & 0.89 & - \\
        SeedFormer\cite{zhou2022seedformer} & 0.72 & 0.81 & 0.40 & - & - & - & 0.50 & 0.77 & 1.49 & 0.92 & 0.472 \\ 
        ProxyFormer \cite{li2023proxyformer} & 0.70 & 0.83 & 0.34 & - & - & - & \textbf{0.49} & 0.75 & 1.55 & 0.93 & 0.483 \\
        AdaPoinTr\cite{yu2023adapointr} & 0.62 & 0.69 & 0.33 & 1.33 & 0.68 & 0.33 & \textbf{0.49} & 0.69 & 1.24 & 0.81 & \textbf{0.503}\\
        \hline
        T-CorresNet & \textbf{0.60} & \textbf{0.68} & \textbf{0.32} & \textbf{1.17} & \textbf{0.60} & \textbf{0.27} & 0.50 & \textbf{0.68} & \textbf{1.23} & \textbf{0.80} & 0.485 \\
    \bottomrule[2pt]
    \end{tabular}
    \label{tab:shapenet55}
\end{table}
\begin{table}
    \caption{Quantitative results on the ShapeNet-34 dataset in terms of CD-$l_2$ $\times$ 1000(lower is better) and F-Score@1\%(higher is better). We use the $\overline{\text{CD}}$ to represent the average CD result.}
    \centering
    \begin{tabular}{c|ccccc|ccccc}
    \toprule[2pt]
       Methods  &  \multicolumn{5}{c|}{34 seen categories} & \multicolumn{5}{c}{21 unseen categories} \\
         & CD-S & CD-M & CD-H & $\overline{\text{CD}}$ & F1 & CD-S & CD-M & CD-H & $\overline{\text{CD}}$ & F1 \\  
         \hline
         FoldingNet\cite{yang2018foldingnet} & 1.86 & 1.81 & 3.38 & 2.35 & 0.139 & 2.76 & 2.74 & 5.36 & 3.62 & 0.095 \\
         PCN\cite{yuan2018pcn} & 1.87 & 1.81 & 2.97 & 2.22 & 0.154 & 3.17 & 3.08 & 5.29 & 3.85 & 0.101 \\
         TopNet \cite{tchapmi2019topnet} & 1.77 & 1.61 & 3.54 & 2.31 & 0.171 & 2.62 & 2.43 & 5.44 & 3.50 & 0.121 \\
         PFNet \cite{huang2020pf} & 3.16 & 3.19 & 7.71 & 4.68 & 0.347 & 5.29 & 5.87 & 13.33 & 8.16 & 0.322 \\
         GRNet\cite{xie2020grnet} & 1.26 & 1.39 & 2.57 & 1.74 & 0.251 & 1.85 & 2.25 & 4.87 & 2.99 & 0.216\\
         PoinTr\cite{yu2021pointr} & 0.76 & 1.05 & 1.88 &1.23 & 0.421 & 1.04 & 1.67 & 3.44 & 2.05 & 0.384 \\
         SnowflakeNet \cite{xiang2021snowflakenet} & 0.60 &0.86 & 1.50 & 0.99 & 0.422 & 0.88 & 1.46 & 2.92 & 1.75 & 0.388 \\
         SeedFormer \cite{zhou2022seedformer} & 0.48 & 0.70 & 1.30 & 0.83 & 0.452 & 0.61 & 1.07 & 2.35 & 1.34 & 0.402 \\
         ProxyFormer\cite{li2023proxyformer} & \textbf{0.44} & 0.67 & 1.33 & 0.81 & 0.466 & 0.60 & 1.13 & 2.54 & 1.42 & 0.415 \\
         AdaPoinTr\cite{yu2023adapointr} & 0.48 & \textbf{0.63} & \textbf{1.07} & \textbf{0.73} & \textbf{0.469} & 0.61 & 0.96 & 2.11 & 1.23 & \textbf{0.416}\\
         \hline
         T-CorresNet & 0.48 & \textbf{0.63} & 1.09 & \textbf{0.73} & 0.462 & \textbf{0.58} & \textbf{0.91} & \textbf{1.97} & \textbf{1.15} & 0.409 \\
         \bottomrule[2pt]
    \end{tabular}
    \label{tab:shapenet34}
\end{table}
We conduct experiments on the ShapeNet-55 benchmark to evaluate the ability of our model to handle more diverse 3D incomplete objects which vary from category to category. We report the average CD-$l_2$ distance on three difficulty degrees and the overall CD-$l_2$ and F-Score@1\% as well in \cref{tab:shapenet55}. In order to evaluate our method comprehensively, we choose three categories of objects(table, chair and airplane), which contain more than 2500 samples in the training set. And the other three categories (birdhouse, bag, and keyboard) have less than 80 samples instead. It can be noticed that our T-CorresNet surpasses other methods on all of the 6 categories and establishes the best Avg CD-$l_2$. While our F-Score@1\% can just be close to the best level, this is due to the noise introduced by the spherical template. Although we use the Corres-Pooling mechanism to refine the coarse template by the sampled input, the unknown part is inevitably noised.  

On the ShapeNet-34 benchmark, the generalization ability of the models is strictly tested by the 21 unseen categories. The results are shown in \cref{tab:shapenet34}. As discussed earlier, T-CorresNet can achieve performance close to the best methods in terms of the F-Score@1\%. It is worth mentioning that our T-CorresNet not only achieve the best CD-$l_2$ on both two types of objects, but the generalization ability of our model outperforms the other methods by nearly 7\%. Additionally, T-CorresNet greatly improves the performance on the CD-H level where the partial input only contains 25\% of the unseen objects. Overall, the design of the Gaussian spherical template and the Corres-Pooling greatly improves the generalization ability of our T-CorresNet.
\subsection{Ablation Study}
\begin{table}[t]
\caption{Ablation Study on the PCN dataset in terms of the critical components of T-CorresNet. We mainly divide our key novelty into the spherical template guide(Template Guide) and the Corres-Pooling query generation strategy (Corres-Pooling). We report CD-$l_1$ and F-Score@1\% to evaluate the performance.}
    \centering
    \begin{tabular}{c|cc|cc}
    \toprule[2pt]
       \makebox[0.2\textwidth][c]{Methods} & \makebox[0.2\textwidth][c]{Template Guide} & \makebox[0.2\textwidth][c]{Corres-Pooling} & \makebox[0.15\textwidth][c]{CD-$l_1$} & \makebox[0.15\textwidth][c]{F1}  \\
       \hline
       baseline &{} & {} & 7.65 & 0.788 \\
       A & $\surd$ & & 7.42 & 0.803 \\
       B & & $\surd$ & 6.58 & 0.841 \\
       \hline
       T-CorresNet & $\surd$& $\surd$& \textbf{6.47}& \textbf{0.845} \\
    \bottomrule[2pt]
    \end{tabular}
    \label{tab:ablation}
\end{table}
In this section, we conduct a detailed ablation study on the critical components of our T-CorresNet, in order to evaluate the effectiveness of our designs. The results are illustrated in \cref{tab:ablation}. The baseline model is constructed upon the modified PoinTr, where we use the fused point proxies instead of the point features as the input of the transformer encoder. The critical components of our T-CorresNet can be summarized as the spherical template guide and the Corres-Pooling query generation strategy, therefore we perform the evaluation through excluding one of these components each time. Specifically, in method A, the coarse complete template is concatenated with the global feature directly to generate the query tokens and it is fed into the FoldingNet without any post-processing. And in method B, the spherical template is removed from the input pipeline. The point proxies extracted from the partial input are directly fed into the encoder to generate the coarse template. The coarse template is then fed into the Corres-Pooling query generation module together with the sampled input to generate the fine template and the dynamic query tokens. Additionally, following the baseline, the standard sphere is removed from value vectors and the point proxies from the local input are directly used as the value vectors.

From the results, We can see that the Gaussian spherical template guide definitely does good to the point cloud completion task and improves the two metrics to some extent. Although the introduction of the template inevitably brings some uncontrollable noise, the template contributes to a certain improvement in the performance of the model. When we exclude the template guide from T-CorresNet, it can be noticed that the Corres-Pooling module further improves the performance by a large degree. It turns out that the spherical template guide and the Corres-Pooling query generation strategy both are effective to the point cloud completion task in our T-CorresNet.

\section{Conclusion}
In this paper, we propose a novel T-CorresNet for the point cloud completion task. In T-CorresNet, a spherical template is used to guide the generation of the coarse complete point cloud which can enlarge the input space and grant the network a better global understanding of the model shape. And a Corres-Pooling query generation strategy is proposed to generate the dynamic and effective query tokens, maintaining the raw information and strengthening the ability to predict point proxies at the same time. The effectiveness of T-CorresNet is evaluated on several challenging benchmarks and our method outperforms the state-of-the-art methods. In future work, we will further explore the design of the spherical template and try to maintain more geometric details during the up-sampling process, which might be the main bottleneck of current point cloud completion methods.  

\par\vfill\par

%
%
\bibliographystyle{splncs04}
\bibliography{main}
\end{document}